\pdfoutput=1
\documentclass[pdflatex,sn-mathphys-num]{sn-jnl}
\makeatletter
\def\ps@headings{%
 \def\@oddhead{\hfil\thepage\hfil}%
 \def\@evenhead{\hfil\thepage\hfil}}
\makeatother

\usepackage{graphicx}%
\usepackage{multirow}%
\usepackage{amsmath,amssymb,amsfonts}%
\usepackage{amsthm}%
\usepackage{mathrsfs}%
\usepackage[title]{appendix}%
\usepackage{xcolor}%
\usepackage{textcomp}%
\usepackage{manyfoot}%
\usepackage{booktabs}%
\usepackage{algorithm}%
\usepackage{algorithmicx}%
\usepackage{algpseudocode}%
\usepackage{listings}%

\theoremstyle{thmstyleone}%
\theoremstyle{thmstyletwo}%

\theoremstyle{thmstylethree}%

\begin{document}

\title[Feature-based morphological analysis of shape graph data]{Feature-based morphological analysis of shape graph data}

\author[1]{\fnm{Murad} \sur{Hossen}}\email{mhossen2@cougarnet.uh.edu}

\author[2]{\fnm{Demetrio} \sur{Labate}}\email{dlabate@uh.edu}
\equalcont{These authors contributed equally to this work.}

\author[3]{\fnm{Nicolas} \sur{Charon}}\email{ncharon@central.uh.edu}
\equalcont{These authors contributed equally to this work.}

\affil[1,2,3]{\orgdiv{Department of Mathematics}, \orgname{University of Houston}, \city{Houston},  \state{Texas}, \country{USA}}

\abstract{This paper introduces and demonstrates a computational pipeline for the statistical analysis of shape graph datasets, namely geometric networks embedded in 2D or 3D spaces. Unlike traditional abstract graphs, our purpose is not only to retrieve and distinguish variations in the connectivity structure of the data but also geometric differences of the network branches. Our proposed approach relies on the extraction of a specifically curated and explicit set of topological, geometric and directional features, designed to satisfy key invariance properties. We leverage the resulting feature representation for tasks such as group comparison, clustering and classification on cohorts of shape graphs. The effectiveness of this representation is evaluated on several real-world datasets including urban road/street networks, neuronal traces and astrocyte imaging. These results are benchmarked against several alternative methods, both feature-based and not.}

\keywords{Shape graph analysis, morphological features, clustering, classification, urban road networks, neuron traces, astrocyte morphology}



\maketitle

\section{Introduction}\label{sec1}
Complex network structures are ubiquitous in both natural systems and engineered environments. From the multiscale ramified branching of neuronal dendrites and other cellular architectures, to the arrangement of blood vessels or lung airways in the human anatomy or root systems in plants, to the sprawling layout of urban road networks and power grids, these structures define the functionality and efficiency of the systems they represent. Unlike standard mathematical graphs that are entirely described by the connectivity information, as for e.g. social or communication networks, many of these examples rather involve graphs embedded in 2D or 3D physical space with the connection between two nodes being itself a geometric curve whose shape may be an equally relevant property of the object. This type of structure is typically referred to as a \textit{spatial network} \cite{barthelemy2011spatial} in the field of network theory, but has also been an important and longstanding subject of interest in the area of statistical shape analysis, where several recent works \cite{guo2020representations,sukurdeep2022new,bal2024statistical} have used the term \textit{shape graph} to designate such geometric objects. One important specificity of this line of work on shape graphs is the emphasis on the need for a combined analysis of the underlying branching structure (i.e., the graph topology) together with the geometric characteristics of the branches themselves, thus placing shape graphs at the intersection between topology and geometry.


Our primary interest in this work lies in population morphological analysis, that is, in solving problems such as clustering or classification on ensembles of shape graphs. Although being basic standard tasks in statistics and machine learning, these tasks become much more challenging when considering data points that are shape graphs rather than usual Euclidean vectors. Indeed, the unique challenge in this case is to deal with the variable and inconsistent topology from one data point to another, combined with the set of geometric invariances (e.g., to rigid transformation or parameterization) that are required to properly compare such objects. 

One of the predominant lines of work towards that goal has been to rely on the definition of a proper metric notion (often a Riemannian metric) on the space of shape graphs from which one can derive a number of statistical analysis tools. This is the case, for instance, of frameworks such as Gromov-Hausdorff, its extension known as the Gromov-Wasserstein metric \cite{memoli2011gromov} and other variants which have been applied to problems similar to this paper \cite{peyre2016gromov,titouan2019optimal,clark2025generalized}. Other related works have alternatively considered extensions of the elastic \cite{guo2020representations,sukurdeep2022new,bal2024statistical} or diffeomorphic metric setting \cite{antonsanti2021partial} from the field of Riemannian shape analysis. One of the known pitfalls of all such approaches, however, is their relatively high computational footprint, as the estimation of just a single distance/geodesic typically involves solving an intricate high-dimensional (often non-convex) optimization problem. This can be a daunting challenge in problems such as clustering of large datasets, which require computations of pairwise distances between points.    

A second main body of work has instead proposed relying on the extraction of features and carrying out statistical analysis tasks in feature space. This approach is usually applied to standard (topological) graphs where  features such as average node degree, assortativity, and spectral entropy are well-established to compare the structure of different networks. However, for shape graphs (or spatial networks), the current state-of-the-art tends to be more scattered, with different works having proposed limited numbers of features tailored to a particular data application, such as transport \cite{barthelemy2011spatial}, geographic \cite{boeing2019urban}, neuronal \cite{Bijari2021,kayasandik2018} or arterial \cite{chen2019quantification} networks. In contrast to such explicitly engineered features, many recent methods are rather attempting to estimate the feature embedding itself, usually via neural network architectures adapted to the structure of shape graph data. Such methods include, most notably, frameworks derived from graph neural networks \cite{wang2019dynamic,kanari2024deep}, the PointNet model of \cite{qi2017pointnet} (designed for 2D and 3D point clouds) and its variants, or other geometric deep learning architectures \cite{bronstein2017geometric,hartman2023varigrad,hartman2025svarm}. Nevertheless, there are several outstanding challenges for the applicability of such models to many real-world datasets, such as those we consider in this work. One major challenge is the size of the dataset. In many biomedical datasets, including the datasets we consider in this work, the size of the data ranges from a few hundred to a few thousands at most, which is insufficient to train the learning models computed in \cite{wang2019dynamic} or \cite{qi2017pointnet} using tens of thousands of samples or more. Another more fundamental obstacle is the difficulty to incorporate all geometric invariances into neural network architectures, in particular invariances to shape graph parametrization, as evidenced empirically in \cite{hartman2023varigrad}. Last but not least, feature representation obtained from neural networks is very difficult or impossible to interpret qualitatively, which is a major limitation to its applications in the life sciences.           

\begin{figure}
    \centering
    \includegraphics[width=1.0\textwidth]{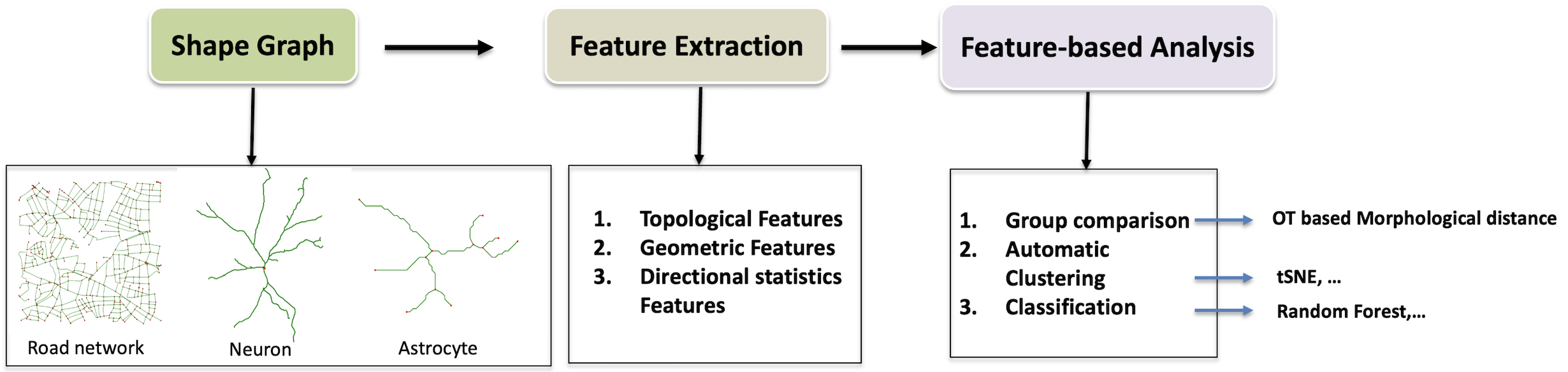}
    \caption{Proposed general data analysis pipeline.} \label{fig.summary}
\end{figure}

To address the limitations of existing methods in processing complex, real-world network structures, we introduce a framework that constructs explicit features on shape graphs and leverages the corresponding feature representation for classification and clustering. Our approach focuses on a comprehensive yet compact set of interpretable features that capture essential topological and geometric characteristics of shape graph datasets. Crucially, our feature selection approach is designed to rigorously enforce fundamental geometric invariances - such as rigid transformations and variations in branch sampling or orientation - thereby eliminating the need for prior registration or resampling across datasets. We benchmark our pipeline (Figure \ref{fig.summary}) against both feature-based and non-feature-based state-of-the-art methods across several real-world datasets on clustering and classification problems. To ensure reproducibility, our Python implementation is publicly available on GitHub :link will show upon acceptance of journal.

\section{Shape Graph}   \label{sec.sg}
Shape graphs are becoming an increasingly common type of structure across various areas of data science. Unlike usual mathematical networks or graphs that encode connectivity as abstract edges between the different nodes, the key specificity of shape graphs is the morphological nature of these connections themselves, which constitute branches from a real object such as a neuron, road network, root system, etc. Importantly, in most applications, one is interested not only in the underlying branching structure (i.e., the graph topology), but also in the geometric characteristics of individual branches, thus placing shape graphs at the intersection between topology and geometry. Several recent works have proposed mathematical frameworks for shape graph representation and analysis with particular focus on shape spaces \cite{guo2020representations,sukurdeep2022new}. Those works typically define a shape graph as a reunion of continuous embedded curves (the branches) connected to each other via some specified adjacency relationships described by a topological graph, modulo a set of geometric invariants (e.g., parametrization$\ldots$). 

In this paper, we adopt a similar approach by directly focusing on the case of discrete shape graphs embedded in some Euclidean space $\mathbb{R}^d$ with $d$ being $2$ or $3$. This means that each branch will be viewed as a discrete curve embedded in $\mathbb{R}^d$. Mathematically, we thus model a shape graph as an unoriented attributed graph, that is, a triple $\mathcal{G} = (V,E,B)$ where $V=\{v_1,\ldots,v_{M}\}$ with $v_i \in \mathbb{R}^d$ is the set of nodes of the graph, $E=\{e_1,\ldots,e_{N}\}$ is the edge set in which each $e \in E$ will be encoded as a pair $e = (i,j)$ where $i,j$ are two distinct node indices $\{1,\ldots,M\}$ (note that we technically do not consider self loops here as these are not encountered in the examples of interest in this paper). Lastly, $B:E \rightarrow X$ is a function defined in $E$ that encodes the geometric branch associated with each edge. It takes values in the space $X$ of all discrete curves of $\mathbb{R}^d$ with initial and terminal points coinciding with the corresponding branch nodes, i.e. for each $e=(i,j) \in E$, $B(e)$ is an ordered list of points that we will denote by $B(e) = (x^e_1,x^e_2,\ldots,x^e_{n_e})\in \mathbb{R}^{n_e \times d}$ with the boundary constraints that $x^e_1 = v_{i}$ and $x^e_{n_e} = v_{j}$. For any edge $e=(i,j) \in E$ and the corresponding branch $B(e) = (x^e_1,x^e_2,\ldots,x^e_{n_e})$, we compute the {\bf edge length} as
\begin{equation}  \label{eq.elength}
  \ell_e  = \| x_{n_e}^e - x_1^e \|_2 = \|v_j - v_i\|_2 
\end{equation}
and the {\bf branch length} as
\begin{equation}  \label{eq.blength}
s_e = \sum_{k=1}^{n_e-1} \| x_{k+1}^e - x_k^e \|_2
\end{equation}

\begin{figure}[htbp]
    \centering
    \includegraphics[width=0.8\textwidth]{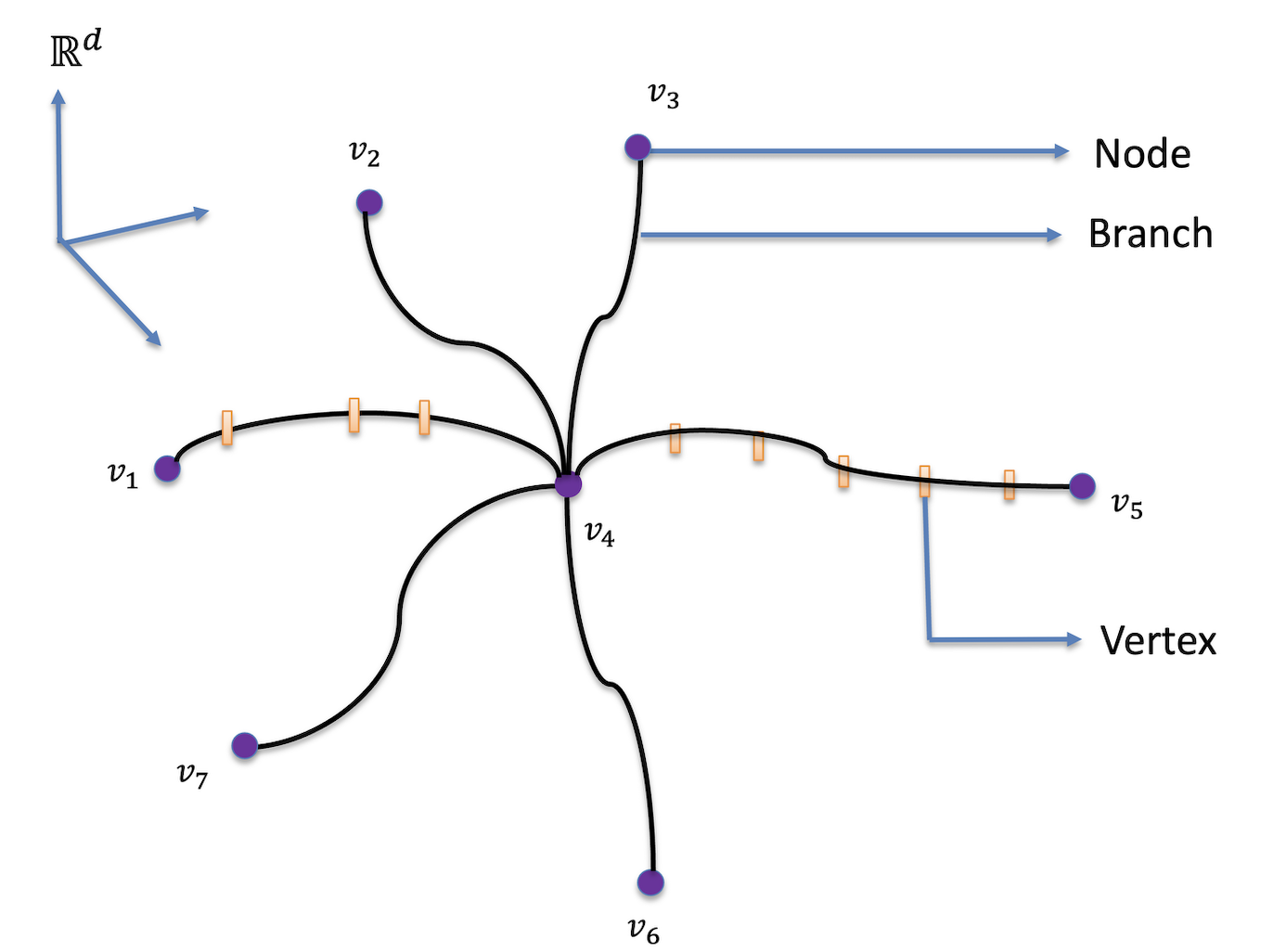}
    \caption{Visualization of Shape Graph}
    \label{fig:shape_graph_illustration}
\end{figure}

An illustration of the different components of a shape graph is shown in Figure \ref{fig:shape_graph_illustration}. We note that, due to the ordering of the initial and final vertices, it technically imposes an orientation to the branch; this will be resolved in the next section by ensuring that the selected shape graph features are independent of the direction of each branch. Our construction also allows the number of points/vertices $n_e$ to depend on the branch itself, which is common in real data where sampling across different branches may vary significantly. Still, one of the fundamental invariance that we intend to achieve with our proposed framework is robustness to branch resampling, which is the discrete equivalent of the reparametrization invariance for the aforementioned continuous shape graph models. In practice, this would ensure that if the vertices of a branch are interpreted as samples from an idealized continuous curve, our analysis should not depend significantly on the sampling pattern being used.  

\section{Invariant features on shape graphs}    
\label{sec.sgfeat}
In this section, we provide the list of invariant features we extract from a shape graph, each accompanied with a brief description, including computational aspects when relevant. The list is organized into 3 groups: topological, geometric, and directional statistics features. Adopting the notation introduced in Section~\ref{sec.sg}, we denote a shape graph as $\mathcal{G} = (V, E, B)$ where $(V,E,B)$ are the sets of edges, nodes, and branches, respectively. 

We used the implementations available in the Python package NetworkX \cite{networkx_2025} to calculate Mean Betweenness Centrality, Algebraic Connectivity, Assortativity, Graph Diameter and Average Shortest Path Length. To compute the Fractal Dimension, we have adapted the implementation from the PoreSpy library \cite{Gostick2019}.

\subsection{\textbf{Topological Features}}
We start with the {\it topological} features which describe those shape graph features that do not depend on the branch geometry but only on the vertex and edge structure $(V,E)$.

\textbf{Number of Edges:} 
The number of edges in a shape graph $\mathcal{G}$ is the cardinality of its edge set $E$ which we denote as $|E|$. Note that this number is also the cardinality of the branch set $B$.
\vskip1ex


\textbf{Mean Betweenness Centrality:} This feature quantifies the importance of nodes in facilitating connectivity \cite{boeing2019urban}.
 For a shape graph \(\mathcal{G}\), the betweenness centrality of a node \(v \in V\) is: 
\begin{equation*}
c_B(v) = \sum_{s \neq v \neq t \in V} \frac{\sigma_{st}(v)}{\sigma_{st}},
\end{equation*}
where \(\sigma_{st}\) is the number of shortest paths between nodes \(s\) and \(t\), computed based on the corresponding edge lengths as defined by~\eqref{eq.elength},
and \(\sigma_{st}(v)\) is the number of those paths that pass through \(v\). Then the mean betweenness centrality over the whole graph is defined as \(c_{B} = \frac{1}{|V|} \sum_{v \in V} c_B(v)\).



\textbf{Spectral Entropy:}  
Spectral entropy measures the complexity of a graph’s weighted Laplacian matrix \cite{bush2015morphological}. For a shape graph $\mathcal{G}$ with $M$ nodes, the weighted Laplacian matrix of the graph is the $M \times M$  symmetric matrix \(L = D - A\), where \(A\) is the adjacency matrix with entries $A_{ij} = \ell_e$ when $e = (i, j) \in E$ and $0$ otherwise, and \(D\) is the diagonal degree matrix with entries \(D_{ii} = \sum_{j} A_{ij}\). The spectral entropy is then defined as  
\begin{equation}
H{} = -\sum_{k} p_k \log p_k, \quad p_k = \frac{\lambda_k}{\sum_m \lambda_m}, ,
\end{equation}
where \(\lambda_k\) are the eigenvalues of \(L\).


\textbf{Algebraic Connectivity:}  
Algebraic connectivity measures the strength of the graph’s connectivity \cite{west2001introduction}. For a shape graph $\mathcal{G}$, the algebraic connectivity is the second smallest eigenvalue \(\lambda_{2}\) of its weighted Laplacian matrix \(L = D - A\) defined above.

\textbf{Assortativity:}
Assortativity measures the tendency of the graph nodes to connect with others with similar weighted degrees \cite{Newman2003}. We here use the previous notion of weighted degree given by the edge length $\ell_e$. The weighted degree, or node strength, of a node $u \in V$ is then
\[
s_u \;=\; \sum_{\substack{e = (u,\cdot) \in E}} \ell_e,
\]
which is the sum of the lengths of all edges incident to $u$. 

Now, for any edge $e = (i,j)$, with corresponding nodes $v_i,v_j$, we set $x_e = s_{v_i}$ and $y_e = s_{v_j}$.  The length-weighted assortativity coefficient $r$ is the Pearson correlation of the pairs $(x_e, y_e)$, that is,
\begin{equation}
r \;=\;
\frac{
\frac{1}{|E|}\displaystyle\sum_{e \in E} x_e y_e
-
\left(
\frac{1}{|E|}\displaystyle\sum_{e \in E} \frac{x_e + y_e}{2}
\right)^2
}{
\frac{1}{|E|}\displaystyle\sum_{e \in E} \frac{x_e^2 + y_e^2}{2}
-
\left(
\frac{1}{|E|}\displaystyle\sum_{e \in E} \frac{x_e + y_e}{2}
\right)^2
}
\end{equation}

\textbf{Graph Diameter:} 
The graph diameter measures the maximum shortest path length between any pair of vertices, reflecting the spatial extent of the graph \cite{boeing2017osmnxnewmethods}. It is given specifically by:
\begin{equation}
\text{diam} = \max_{u,v \in V} d(u,v),
\end{equation}
where \(d(u,v)\) is the shortest path length between the vertices \(u\) and \(v\), where the edge weights are again the lengths $\ell_e$ from~\eqref{eq.elength}.  

\subsection{\textbf{Geometric Features}}

\hspace{\parindent} \textbf{Average Branch Length:}
For a shape graph $\mathcal{G}$, the average branch length is the mean of the branch lengths $s_e$ over all edges $e \in E$, where $s_e$ is given by~\eqref{eq.blength}, i.e.
\begin{equation}
    \bar{s} = \frac{1}{|E|} \sum_{e \in E} s_e.
\end{equation}

\vskip1ex

\textbf{Branch Density:}
The branch density of a shape graph $\mathcal{G}$, denoted $ds_{B}$, is defined as its total branch length divided by the volume of the convex hull enclosing all points in the graph, i.e. the convex hull of all points $x_1^e, x_2^e, \dots, x_{n_e}^e \in B(e)$ for all $e \in E$, that we write $\text{Conv}(B)$. Denoting as $\lambda^d(\text{Conv}(B))$ the Lebesgue measure of that set, that is, its area (if $d=2$) or volume (if $d=3$), the branch density is then given by:
\begin{equation}
ds_B = \frac{\sum_{e \in E} s_e}{\lambda^d(\text{Conv}(B))}.
\end{equation}

\textbf{Bending Energy:}  
The average bending energy of a shape graph $\mathcal{G}$ quantifies the curvature complexity of its branches. We recall that, for a smooth curve \(\gamma\) parameterized by arc length \(s\), the  bending energy of the corresponding elastic rod is given by the line integral of the squared curvature \(\kappa\) i.e.:  
\begin{equation}
\label{eq.bending_curve}
\mathcal{E}(\gamma) = \frac{1}{L} \int_{\gamma} \kappa(s)^2 \, ds,
\end{equation} 
where $L$ is the curve’s total length \(L = \int_{\gamma} ds\). The average bending energy of $\mathcal{G}$ can then obtained as the average of the bending energies of its branch curves, in other words:
\begin{equation}
    \mathcal{E}(\mathcal{G}) = \frac{1}{N} \sum_{e \in E} \mathcal{E}(B(e))
\end{equation}

As each branch curve $B(e)$ is here discrete and represented by the sequence of points \(\{x_1^e, x_2^e, \dots, x_{n_e}^e\} \subset \mathbb{R}^d\), the curvature and integral in \eqref{eq.bending_curve} are here understood in the discrete geometry sense. Specifically, we use the approximate curvature at vertex $x_k^e$, for $k=2,\ldots,n_e -1$, given by the centered scheme $\kappa(x_{k}^e) \approx 2\frac{u_{k}^e - u_{k-1}^e}{s_k^e + s_{k-1}^e}$,
where $s_k^e = \|x^{e}_{k+1} - x_{k}^e\|_2$ and $u_{k}^e = (x^{e}_{k+1} - x_{k}^e)/s_k^e$.

\textbf{Average Shortest Path Length:}  
The average shortest path length measures the efficiency of connectivity between nodes \cite{boeing2017osmnxnewmethods}. For a shape graph $\mathcal{G}$ with \(n = |V|\) nodes, the average shortest path length is:  
\begin{equation}
\text{APL} = \frac{1}{n (n - 1)} \sum_{u \neq v \in V} d(u, v),
\end{equation} 
where \(d(u, v)\) is here the shortest path length between nodes \(u\) and \(v\), weighted by the branch lengths $s_e$ given by~\eqref{eq.blength}.

\textbf{Average Circuity:} 
The average circuity in a shape graph measures the tortuosity over all of its branches \cite{boeing2019urban} which is defined as the ratio between the total branch length and the total edge length, that is
\begin{equation}
\varsigma = \frac{\sum_{e \in E} s_e}{\sum_{e \in E} \ell_e},
\end{equation} 
with $\ell_e$ and $s_e$ given by~\eqref{eq.elength} and \eqref{eq.blength}, respectively.



\begin{figure}[htbp]
    \centering
    \includegraphics[width=0.8\textwidth]{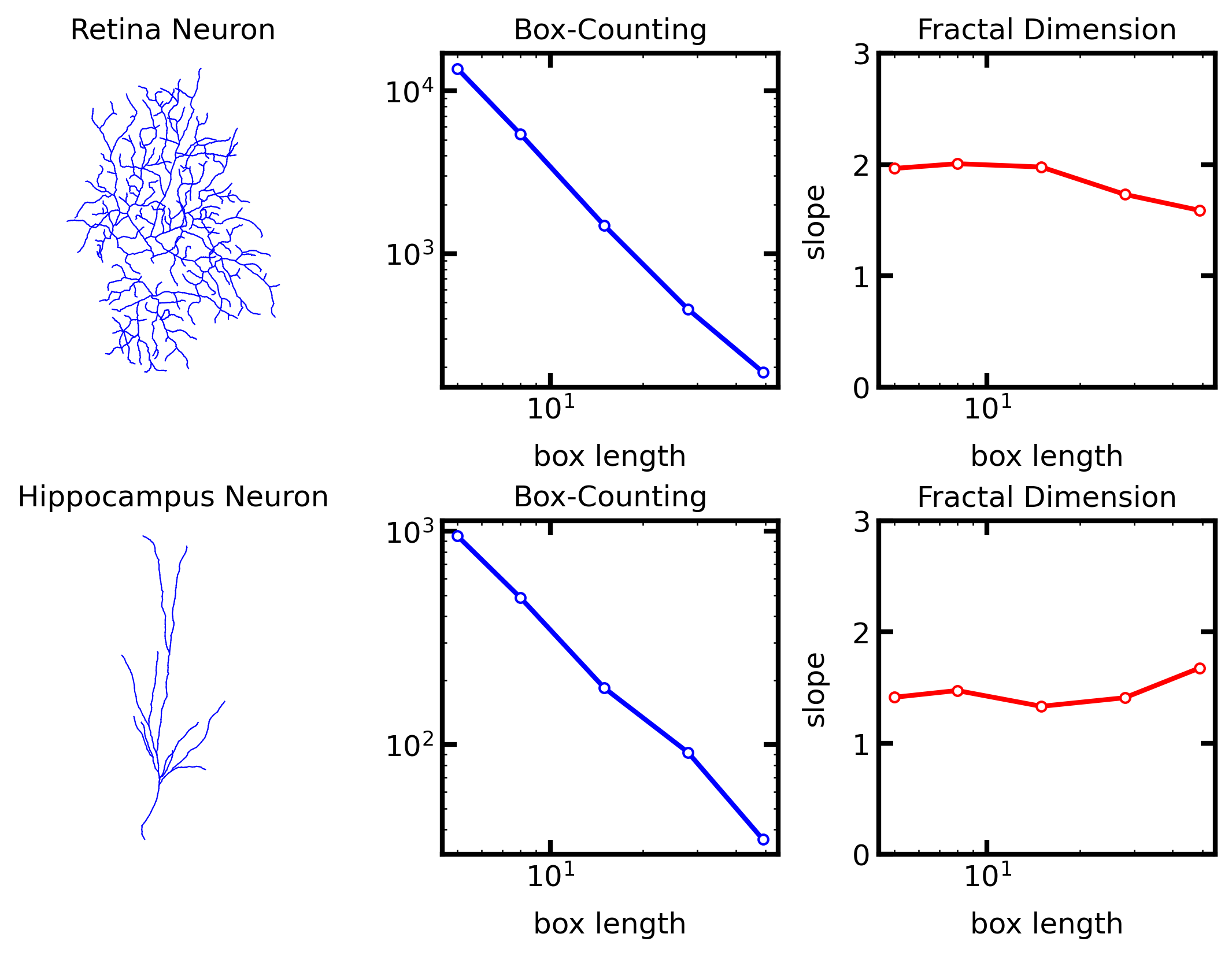}
    \caption{Fractal dimension calculation using neuronal reconstruction images. The input image is first processed to extract the shape graph (including the extraction of the graph branches), next the box-counting log-log plots is computed, finally the fractal dimension is obtained from the slope of the box-counting plot.}
    \label{fig:neuron_fd_calculation}
\end{figure}

\textbf{Fractal Dimension:}  
The fractal dimension of a set quantifies the complexity of its spatial pattern \cite{mandelbrot1967}. The fractal dimension of a shape graph $\mathcal{G}$ is computed using the box-counting method implemented in the PoreSpy library, where a grid of boxes of side length \(D\) is overlaid on the branches of the graph (embedded in the appropriate Euclidean space, $\mathbb{R}^2$ for planar images, $\mathbb{R}^3$ for volumetric images), and one records the number of boxes \(N(D)\) intersecting a branch of the shape graph. The relationship is modeled as:  
\begin{equation}
N(D) \propto \frac{1}{D^F},
\end{equation}
which, by taking the logarithm, yields:  
\[
F = \lim_{D \to 0} \frac{-\log N(D)}{\log D}.
\]  
In practice, \(F\) is estimated as the negative slope of the linear region in a log-log plot of \(N(D)\) versus \(D\), using a range of box sizes. This is illustrated in Figure~\ref{fig:neuron_fd_calculation}.

\subsection{\textbf{Directional Statistics Features}}
Finally, we introduce a set of shape graph features derived from the concept of directional statistics \cite{Fisher1993,Mardia2000} that are designed to capture the overall distribution of the directions of the shape graph curves. 

\subsubsection{Directional histogram of a shape graph}
Let us start by briefly describing the directional distribution associated to a given shape graph $\mathcal{G} = (V, E, B)$. Using the same notations as above, for each edge $e \in E$, and the corresponding polygonal branch curve $B(e) = (x_1^e, x_2^e, \ldots, x_{n_e}^e)$, one computes, for $k=1,\ldots,n_e-1$, the unit direction vectors $u_k^e \doteq (x_{k+1}^e - x_k^e)/\|x_{k+1}^e - x_k^e\|_2 \in \mathbb{S}^{d-1}$ of the different segments together with their lengths $s_k^e \doteq \|x_{k+1}^e - x_k^e\|_2$. Then, it is common to consider the sum of Dirac distribution $\sum_{e \in E} \sum_{k=1}^{n_e-1} s_k^e \, . \, \delta_{u_k^e}$ which is a positive measure on the circle $\mathbb{S}^{1}$ (when $d=2$) or the sphere $\mathbb{S}^{2}$ (for $d=3$), sometimes referred to as the {\it length measure} \cite{charon2021length}. It represents the oriented directional distribution of the whole collection of branch curves.

\begin{figure}[htbp]  
    \centering  
    \includegraphics[width=0.9\textwidth]{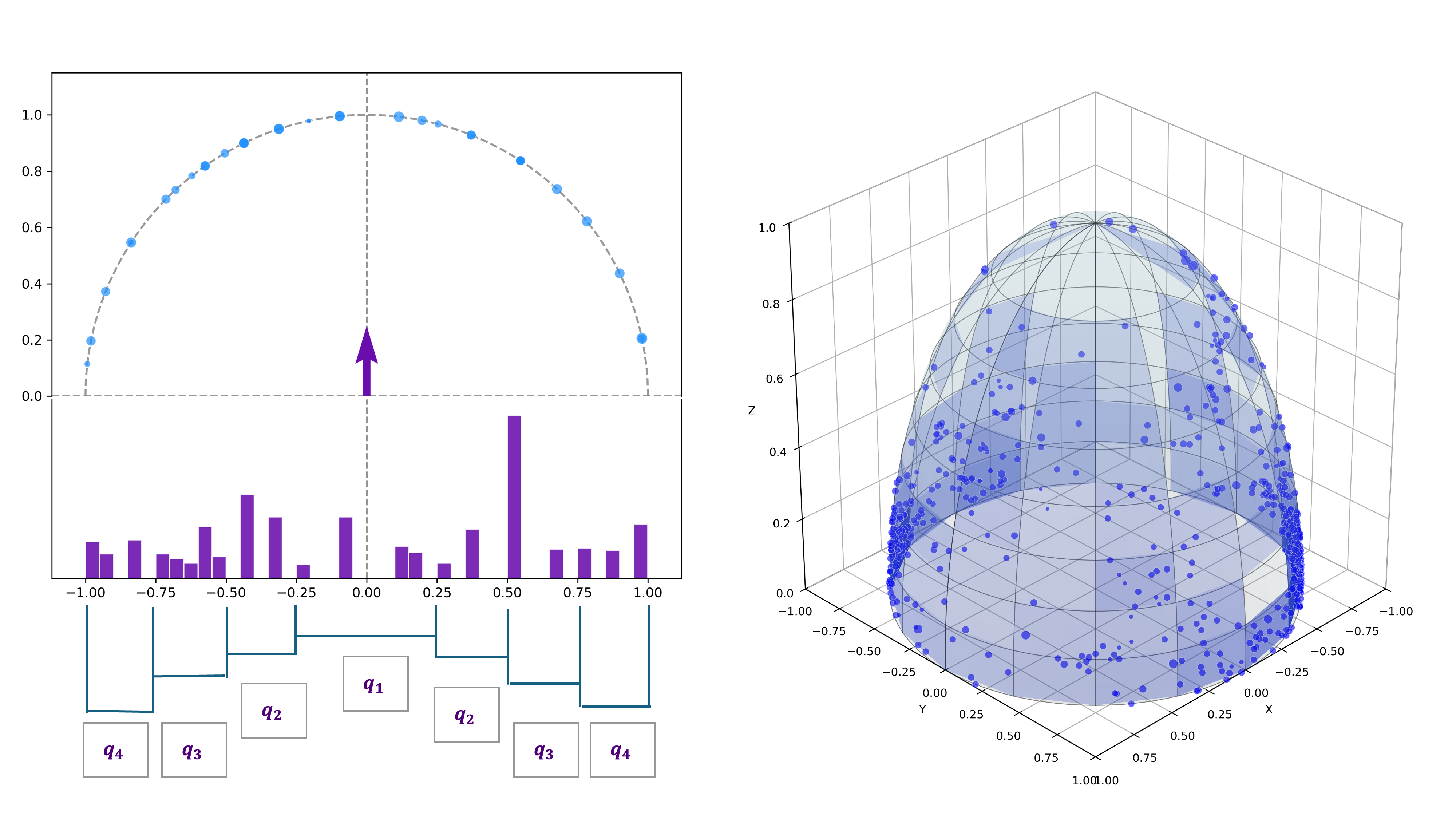}  
    \caption{Visualization of directional histograms of 2D (left) and 3D (right) shape graphs. The original sum of Dirac distributions are plotted as markers (with size proportional to the weight) on the half circle and half sphere, respectively.}  
    \label{fig:directional Statistics}  
\end{figure}

In this work, however, we are interested in shape graphs that do not necessarily carry a natural or consistent notion of orientation along their respective branches. For this reason, we consider the less common "unoriented" variant of the above directional distribution, in which directional components $u_i^e$ are instead viewed as elements of the real projective space $\mathbb{RP}^{d-1}$, i.e., the quotient of $\mathbb{S}^{d-1}$ by the action of the orientation group $\{\pm \text{Id}\}$. We thus introduce the following \textit{unoriented directional distribution} on $\mathbb{RP}^{d-1}$, defined by:
\begin{equation}
    \label{eq:def_direct_dist}
    \mu_{\mathcal{G}} = \sum_{e \in E} \sum_{k=1}^{n_e-1} s_k^e \, . \, \delta_{u_k^e}
\end{equation}
with the convention that two opposite directions $u$ and $-u$ are identified with each other. Specifically, for $d=2$, we represent each $u_i^e \in \mathbb{RP}^{1}$ by the unit vector's unoriented angle $\theta_i^e \in [0,\pi)$. For $d=3$, we can view $\mathbb{RP}^{2}$ as the union of the upper unit open half sphere and a copy of $\mathbb{RP}^{1}$ (accounting for the equatorial circle modulo $\{\pm \text{Id}\}$). In both cases, the distribution $\mu_\mathcal{G}$ can then be converted into a directional histogram by binning the domain $\mathbb{RP}^{d-1}$ and assigning to each bin the sum of the weights $s_k^e$ for the $u_k^e$'s that fall into that bin, as Figure illustrated by~\ref{fig:directional Statistics}.   

\subsubsection{Directional features}
\label{ssec:direc_features}
We note that the measure $\mu_\mathcal{G}$ (or its corresponding histogram) is translation invariant as well as robust to branch resampling and orientation, but it is not rotation-invariant. In what follows, we describe a specific set of features computed from $\mu_\mathcal{G}$ that also satisfy the rotation invariance requirement.

\vskip1ex
\textbf{Standard deviation:} The definition and computation of the mean, variance and standard deviation for distributions on projective spaces is not as straightforward as in the Euclidean setting. However, in the case of $\mathbb{RP}^1$ (i.e., $d=2$), it can be simply reduced to the more standard framework of circular statistics developed in details in \cite{Fisher1993}. Indeed, one can rely on the doubling angle function $\theta \in [0,\pi) \mapsto 2\theta \in [0,2\pi)$, which maps $\mathbb{RP}^1$ to $\mathbb{S}^1$ isometrically (up to a rescaling factor of $2$). Thus, one can simply introduce $\check{\mu}_{\mathcal{G}} = \sum_{e \in E} \sum_{k=1}^{n_e-1} s_k^e \, . \, \delta_{\check{u}_k^e}$, where $\check{u}^e_k = (\cos(2\theta_k^e),\sin(2\theta_k^e)) \in \mathbb{S}^1$, which is a distribution on the unit circle, and compute its mean $\bar{\theta}$ standard deviation $\sigma$ according to the following formula from \cite{Fisher1993}:
\begin{align}
\label{eq:circular_std}
    \bar{u}&=\frac{\sum_{e,k} s_k^e \, \check{u}_k^e}{\sum_{e,k} s_k^e}, \ \ \bar{\theta} = \text{angle}(\bar{u}) \nonumber\\
    \sigma &= \sqrt{-2\ln \|\bar{u}\|}
\end{align}
from which one obtains the mean unoriented angle $\bar{\theta}/2$ for the unoriented distribution $\mu_{\mathcal{G}}$, together with the same standard deviation $\sigma$. By construction and eq.\eqref{eq:circular_std}, $\sigma$ is invariant to the action of any planar rotation applied to the shape graph, and we thus select $\sigma$ as our first directional feature. 

In the case $d=3$, there is no equivalent direct approach to estimate the mean and standard deviation of a distribution on $\mathbb{RP}^2$. As this computation is more technical, we rather leave its description to the appendix.  

\vskip1ex
\textbf{Quantile vector:} 
As a way to recover a more precise description of the directional histogram, it is also common to introduce its quantiles, measuring the proportion of the distribution's mass lying within specific angular intervals around the mean. In order to obtain a normalized and compact feature representation, we consider the vector of the four quantiles of the distribution that we denote $q=(q_1,q_2,q_3,q_4)$. These are specifically computed as follows. For $d=2$, we first center (i.e., rotate) the distribution so that the unoriented mean angle (computed as above) is set to $\pi/2$. Then $q_1,\ldots,q_4$ are given as the proportion of the total mass for which the unoriented angles lie in the respective intervals $I_1=[3\pi/8, 5\pi/8]$, $I_2=[\pi/4, 3\pi/8] \cup [5\pi/8, 3\pi/4]$, $I_3=[\pi/8, \pi/4] \cup [3\pi/4, 7\pi/8]$ and $I_4 = [0, \pi/8] \cup [7\pi/8, \pi]$. This is illustrated on the left panel of Figure \ref{fig:directional Statistics}. This results in a vector $q$ belonging to the probability simplex of $\mathbb{R}^4$. For $d=3$, we follow a similar process in which the distribution is first centered on the vertical polar direction, and the four quantiles are divided according to polar portions of the upper half sphere of equal area. 

In either case, one obtains a quantile vector $q$ living on the probability simplex of $\mathbb{R}^4$ that is invariant with respect to the orientation as well as any rigid transformation of the underlying shape graph.

\vskip1ex
\textbf{Directional entropy:}  
In addition to the standard deviation and quantiles, another potentially informative feature is the entropy of the histogram of directions which measures the overall isotropy/anisotropy of the underlying shape graph. Denoting by $N$ the chosen number of bins and \(p_k\), $k=1,\ldots,N$, the probability of the k-th bin (i.e., the proportion of the total mass of unoriented directions falling in that bin), the directional entropy of the shape graph is then the usual Shannon entropy of the histogram, that is:  
\begin{equation}
\label{eq:directional_entropy}
H_{w} = -\sum_{k=1}^N p_k \log p_k.
\end{equation}   
In practice, we select \(N = 18\) equally spaced bins of \([0, \pi)\) in the case $d=2$, and $N=N_\alpha \, N_\beta$ equal area bins of the upper half sphere for $d=3$, where $N_\alpha = 8$ and $N_\beta = 16$ are the number of polar and azimuthal angular intervals respectively. Figure \ref{fig:directional Statistics} illustrates both situations.

\vskip1ex
\textbf{Orientation-Order:}  
Another common directional feature is the so-called orientation-order indicator which is directly computed from the above directional entropy according to:  
\[
\phi = 1 - \left( \frac{H_{w}}{H_{\max}} \right)^2,
\]  
where \(H_{\max} = \log N\) is the maximum entropy for a histogram with \(N\) bins. The indicator \(\phi\) ranges from 1 (perfect alignment, \(H_{w} = 0\)) to 0 (uniform distribution, \(H_{w} \approx H_{\max}\)). Although it is redundant with directional entropy, we still include this feature in our approach as it provides an interpretable measure
which is often used in practice.

\section{Feature-based statistical analysis}
In this section, we outline the machine learning methods 
that we have applied to different shape graph datasets in combination with the descriptive features presented in Section~\ref{sec.sgfeat}. We focused on three classical supervised or unsupervised tasks, namely, group comparison, automated clustering and classification. Our numerical results are reported in Section \ref{sec.results}. 

\subsection{Group Comparison}
\label{ssec.group_comp}
As our first problem of interest, we wish to compare qualitatively and quantitatively different known groups within the data, based on the differences of their feature distributions. Hence, we assume that a set of shape graphs is (partitioned into $P$ groups $\{\mathcal{G}_i^{(p)}\}_{i=1,\ldots,n^{(p)}}$. For each shape graph $\mathcal{G}_i^{(p)}$, one computes the $N=19$ features described in Section \ref{sec.sgfeat} whose values are all normalized to the interval $[0,1]$. 

In order to measure proximity or differences between these groups, we adopted a strategy similar to previous work such as \cite{Marini2025}. The feature values of a group are binned into a probability histogram which we denote as $\Pi_f^{(p)}$, for each group $p$ and each feature $f$, and that represents the distribution of the values of feature $f$ in group $p$. The framework of optimal transport (OT) then provides a natural approach to quantify the discrepancy between two probability distributions via the family of Wasserstein distances. We specifically leverage the 1-Wasserstein metric $W_1$, also known as the {\it earth mover's metric} \cite{rubner1998metric}, and its implementation in the POT library \cite{flamary2021pot}, to measure the distance $W_1(\Pi_f^{(p)},\Pi_f^{(p')})$ between the groups $p$ and $p'$ for a given feature $f$. As we illustrate in more detail in Section \ref{sec.results}, the resulting distance matrix $(W_1(\Pi_f^{(p)},\Pi_f^{(p')}))_{p,p'=1,\ldots,P}$ can be used to infer the similarity or dissimilarity of different groups with respect to each particular feature $f$. Importantly, as our proposed features are handcrafted, this leads to interpretable comparative analysis.

In addition to the feature-by-feature approach above, it is useful to have a metric summarizing the group differences across all features. For that, we adopted the notion of morphological distance (MD) introduced in~\cite{Marini2025}.  Rather than directly adding or averaging the previous Wasserstein metrics over the different features $f$, which would not necessarily account for potential redundancies or correlations between certain features, MD aggregates those based on their Spearman correlation matrix $R \in \mathbb{R}^{N \times N}$, whose entries $R_{ij}$ are given as the Spearman correlation coefficients of the features $f_i$ and $f_j$ across all sample shape graphs in all groups. The squared MD between groups $p$ and $p'$ is then:
\begin{align}
\label{eq.aggr_morph_dist}
\text{MD}^2(p, p') &= d^T(\Pi^{(p)}, \Pi^{(p')}) \, R \, d(\Pi^{(p)}, \Pi^{(p')}) \nonumber \\
&-\frac{1}{2N} d^T(\Pi^{(p)}, \Pi^{(p')}) \, |\!R - I_N\!| \, d(\Pi^{(p)}, \Pi^{(p')}),
\end{align}
where $d(\Pi^{(p)}, \Pi^{(p')}) \in \mathbb{R}^{N}$ is the vector of the $1$-Wasserstein distances between $\Pi_f^{(p)}$ and $\Pi_f^{(p')}$ for the different features and $I_{N}$ is the identity matrix. For uncorrelated features (i.e. $R = I_{N}$), \eqref{eq.aggr_morph_dist} 
reduces to the sum of the squared Wasserstein distances. However, for correlated features, the subtracted term diminishes the influence of redundant ones based on their correlation strength. As a byproduct, the information from the matrix $R$ can also be used for the correlation analysis of the proposed features, as we will showcase in Section \ref{sec.results}.

\subsection{Automatic Clustering}
\label{ssec.clustering}
The second problem we considered is clustering, namely the unsupervised identification of relevant clusters in a shape graph set. Unlike in the above section, here we assume that we are simply given a set of shape graphs $\{\mathcal{G}_i\}_{i=1,\ldots,n}$ without any group label. The use of a feature embedding of the data coupled with clustering methods (such as k-means, spectral clustering$\ldots$) applied in the feature space is a common pipeline encountered in many applications.  

We illustrate in Section \ref{sec.results} that our proposed low-dimensional invariant feature representation of shape graphs can be effectively leveraged for  clustering of complex branching structures, including urban road networks and neuronal morphologies. More precisely, our chosen approach for clustering will consist of the following simple steps. As in the above section, we compute the list of $N=19$ features for each shape graph $\mathcal{G}_i$ that we normalize in the same way. We then compute the pairwise Euclidean distances between the corresponding feature vectors of $\mathbb{R}^{N}$. The relationships between the different data points are then mapped and visualized in just two dimensions using t-distributed Stochastic Neighbor Embedding (t-SNE) \cite{hinton2002stochastic}, from which a specified number of clusters can then be extracted based on a standard agglomerative clustering scheme. The quality of the resulting partition of the data is assessed through the Adjusted Rand Index (ARI), which compares the estimated labels against ground-truth categories (e.g., urban map type for road networks or brain regions for neurons).

To validate our proposed clustering pipeline, we will benchmark it against comparable approaches based on either different established feature representations in the literature or other morphological distances. These include, in particular, the family of Gromov--Wasserstein (G-W) distances~\cite{memoli2011gromov} defined between any metric measure spaces. In the specific case of shape graphs, it yields a rigid-invariant distance that is numerically computed based on the implementation from the POT library, by solving the optimization problem
$$\min_T \sum_{i,k=1}^{N_1} \sum_{j,l=1}^{N_2} (D^{1}_{ik}-D^{2}_{jl})^2 T_{ij} T_{kl}$$
where the minimization is over all coupling matrices $T \in \mathbb{R}_+^{N_1 \times N_2}$ (with $N_1$, $N_2$ being the number of vertices in the first and second shape graph) subject to $T \mathbf{1}_{N_2} = \frac{1}{N_1} \mathbf{1}_{N_1}$ and $T^T \mathbf{1}_{N_1} = \frac{1}{N_2} \mathbf{1}_{N_2}$, where $D^1$, $D^2$ are the pairwise distance matrices between the vertices of the first and second graph, respectively. Although the above problem is non-convex, it can be solved via a conditional gradient scheme, as proposed in \cite{titouan2019optimal}. 


\subsection{Classification}
\label{ssec.classif}
Lastly, we turn to the common classification task. In this case, we assume that a training set is available, made of shape graphs $\{\mathcal{G}_i^{(c)}\}_{i=1,\ldots,n^{(c)}}$ divided among known classes $c=1,\ldots,C$. In contrast to the group comparison task in Section \ref{ssec.group_comp}, our purpose here is to derive an explicit classification rule based on the extracted morphological features and to evaluate the classification accuracy using a separate testing set. 

There are many standard classification schemes that can be applied to the feature representation we propose, ranging from linear methods such as support vector machines and linear discriminant analysis to nonlinear ones such as deep learning based approaches. As the purpose of this work is not specifically to compare the choice of classification approach but rather to evaluate the effectiveness and flexibility of our feature model, we will restrict ourselves to a very commonly used family of classifiers derived from random forests \cite{breiman2001random}. Another advantage of random forests, in our context, is that they achieve a good balance between classification power and interpretability, in particular given the quite limited amount of training data available in most applications involving shape graphs.

In the results presented in the next section, we specifically adopted a one-versus-rest strategy, fitting a separate binary classification model on the shape graph features for each class. In situations involving imbalanced classes (such as the astrocyte dataset in Section \ref{ssec.results.astrocyte}), we further employed an oversampling procedure based on an adapted SMOTE algorithm \cite{Wang2021} to balance the different classes. This method generates synthetic sample points distributed closer to the center of the minority class (within a restricted standard deviation interval of $0, \sigma_0/3$) to reduce marginalization during training. We will also benchmark the proposed random forest feature approach against alternative methods, which include morphoGNN~\cite{zhu2023data}, a graph neural network model incorporating triplet loss and cross-entropy loss for the training process, as well as SVarM~\cite{hartman2025svarm}, a recent geometric deep learning model that leverages varifold embedding of shapes and was shown to achieve competitive performance with multiple state-of-the-art deep learning architectures for shape classification.  

To evaluate a classifier performance, we adopted usual metrics for binary classification which we briefly summarized below. 

The \textit{sensitivity} $S$, also known as the true positive rate or recall, is the fraction of accurately identified instances relative to all actual instances in a class, i.e.:
\[
S = \frac{\text{TP}}{\text{TP} + \text{FN}}.
\]
where TP denotes the total number of true positives (correct predictions) and FN the number of false negatives (missed predictions). Second, the \textit{precision} $P$ determines the proportion of correct predictions among all instances labeled as that class, i.e.:
\[
P = \frac{\text{TP}}{\text{TP} + \text{FP}}.
\]
with $\text{FP}$ denoting the number of false positives. In addition, we can also evaluate the \textit{F1 score} that offers a balanced view of the classifier's effectiveness and is given as the harmonic mean of precision and sensitivity:
\[
\text{F1} = 2 \frac{P \cdot S}{P + S} = \frac{2 \cdot \text{TP}}{2 \cdot \text{TP} + \text{FP} + \text{FN}}.
\]
Finally, the classification \textit{accuracy} reflects the overall ratio of correct predictions and is widely used when classes hold equal importance:
\[
\text{Accuracy} = \frac{\text{TP} + \text{TN}}{\text{TP} + \text{FP} + \text{TN} + \text{FN}},
\]
where TN stands for the number of true negatives (correctly excluded instances).

\section{Experimental validation}
\label{sec.results}
We shall now present the evaluation of the feature-based analysis methods described in the previous section on three different real shape graph datasets and benchmark the results against a few alternative approaches from the state-of-the-art.  


\subsection{City road networks}
\subsubsection{Data description}
\begin{figure}[htbp]
  \centering
  \includegraphics[width=0.9\textwidth]{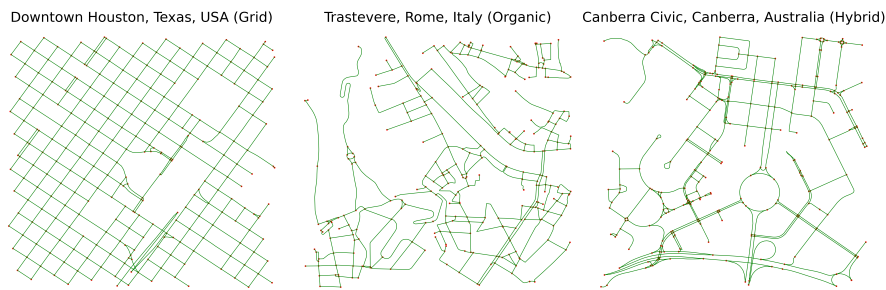}
  \caption{Street networks for Downtown Houston (Grid), Trastevere district in Rome (Organic) and Canberra (Hybrid). Nodes and branches are plotted in red and green, respectively.}
  \label{fig:city_maps_subplot}
\end{figure}

As a first example, we will consider urban street network data extracted via OpenStreetMap, a collaborative open-source mapping project providing high-quality, global geospatial location and shape of streets, buildings, and amenities. 
The dataset was specifically retrieved using the OSMnx library \cite{boeing2017osmnxnewmethods, Boeing2025}, where we selected a total of 100 urban areas of size 800m $\times$ 800m located across seven continents (North America: 24, South America: 8, Europe: 33, Asia: 13, Africa: 9, Australia \& Oceania: 6, Others: 7). These were chosen so as to showcase different types of urban street layouts, which we categorized into "grid" (e.g., Downtown Houston, USA), "organic" (e.g., Trastevere, Rome), and "hybrid" (e.g., Canberra, Australia) based on their spatial patterns of orthogonality, irregularity, or a mixture thereof following similar criteria as in \cite{boeing2019urban, Boeing2025}. Some example of these street maps are shown in Figure \ref{fig:city_maps_subplot}. These different categories emphasize global variety in urban design which are influenced by multiple factors including historical, cultural, or topographical.

\subsubsection{Group comparison and clustering}
\label{ssec.cluster_neuron}
\begin{figure}
    \centering
    \includegraphics[width=0.6\linewidth]{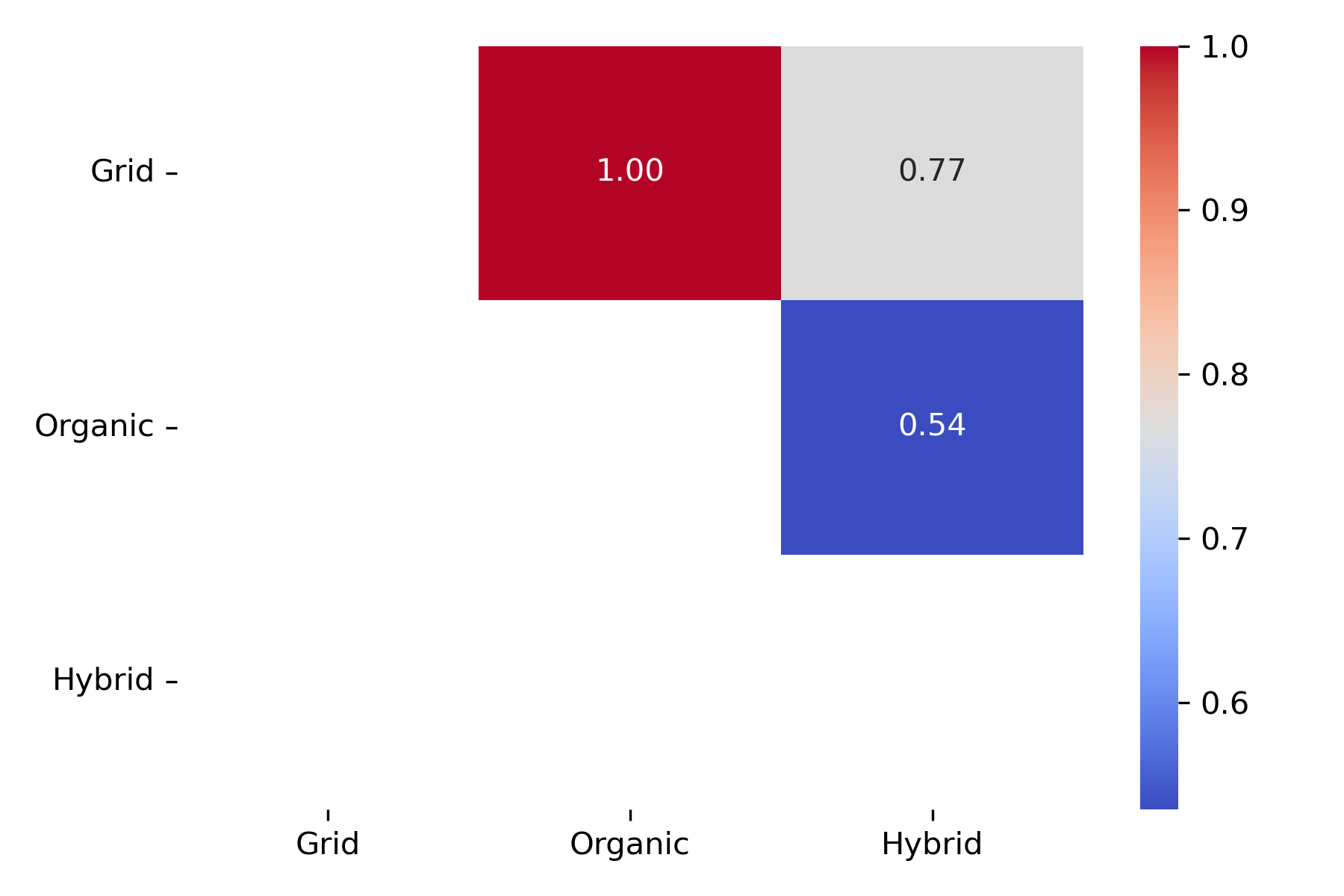}
    \caption{Normalized pairwise morphological distances between the three groups of road networks.}
    \label{fig:Morphological_distance_road}
\end{figure}

We first show in Figure \ref{fig:Morphological_distance_road} the result of the pairwise morphological distances described in Section \ref{ssec.group_comp} between the three different groups of road networks. As one could naturally expect in this case, the maximal distance is between the two extreme groups "Grid" and "Organic" while the "Hybrid" is roughly at comparable distance to the other two groups.

\begin{figure}
    \centering
    \includegraphics[width=1.0\linewidth]{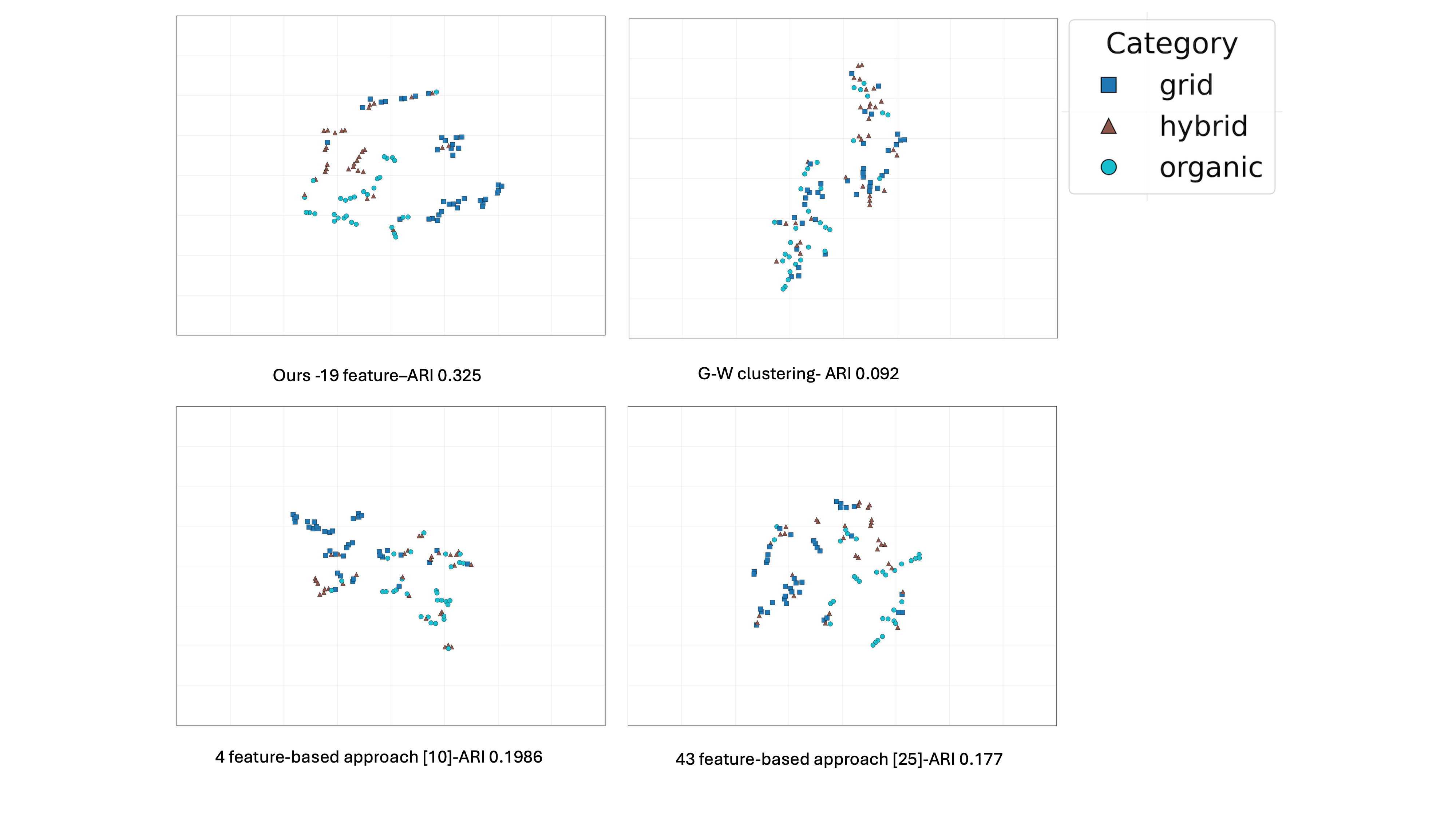}
    \caption{Comparison of clustering obtained from different distance representations, visualized via t-SNE. Adjusted Rand Index (ARI) measures agreement between discovered clusters and ground-truth morphological categories (grid/hybrid/organic). Our 19-feature based approach achieves the highest ARI and shows clearer visual separation between the three different street network categories.}
    \label{fig:tsne_clustering}
\end{figure}

We next examine the clustering approach described in Section \ref{ssec.clustering} applied to this dataset. Specifically, we aim to measure how much the clusters obtained via this unsupervised approach overlap with the underlying assigned labels. We also compare results with the Gromov-Wasserstein distance clustering mentioned in Section \ref{ssec.clustering}, as well as two other feature-based approaches previously considered for road network analysis: one based on the set of 43 OSMnx summary statistics features from \cite{boeing2017osmnxnewmethods}, and the other \cite{boeing2019urban} relying on a set of four specific features (orientation order and average circuity as in our approach, plus median street length and average node degree). For each method, we compute the 2D t-SNE embedding of the data based on the respective Euclidean feature distances. The resulting embeddings are shown in Figure \ref{fig:tsne_clustering}. We can see that no clear cluster pattern seems to emerge in the G-W case or for the approach of \cite{boeing2017osmnxnewmethods}, while the embeddings associated to our proposed feature model or that of \cite{boeing2019urban} appears to show clearer group structure. Applying agglomerative clustering (with a prescribed number of 3 clusters) to the embedded data coordinates, we may further evaluate how those computed clusters overlap with the "ground-truth" categories via the estimated ARI between the cluster labels and the "ground-truth" categories. The ARI values are also reported on Figure \ref{fig:tsne_clustering}, showing a highest score for our approach. This suggests that the proposed set of features is better able to retrieve the main geometric characteristics that distinguish the different types of urban street layout.    

\subsubsection{Feature-based classification of road networks}
\label{ssec.classif_roads}
Lastly, we evaluate the random forest classification approach described in Section \ref{ssec.classif}. We specifically allocated 70\% of the dataset to training and left the remaining 30\% for testing. To reduce bias in sample allocation, we executed 10 random divisions and averaged the findings. The model can successfully anticipate the structural types of road networks, attaining an overall accuracy of 72.33 $\pm$ 7.31. Specific accuracies spanned from 83.33 $\pm$ 10.54 for grid, 70.00 $\pm$ 15.75 for organic, to 60.00 $\pm$ 15.87 for hybrid, indicating higher accuracy for grid patterns with greater variability in the hybrid category.


To gain insight into the features distinguishing capabilities, we ordered them by significance using mean decrease in impurity (MDI), a standard technique for gauging feature relevance in tree-based ensembles like Random Forest. MDI quantifies each feature's contribution by calculating the total decrease in node impurity averaged over all trees when splits on that feature occur. We averaged importances over the 10 runs, including standard deviations.
In the road network data, leading features included circuity, number of edges, orientation order, directional entropy, bending energy. These results affirm the features' effectiveness in differentiation, highlighting aspects like circuity and orientation order as key discriminators between planned, organic, and hybrid urban layouts.
\begin{table}[ht]
\centering
\begin{tabular}{l r}
\hline\hline
Model & Overall Accuracy \\ [0.5ex]
\hline
Random Forest(Ours) & 72.33 $\pm$ 7.31 \\
morphoGNN~\cite{zhu2023data} & 60.00 $\pm$ 4.46 \\
SVarM~\cite{hartman2025svarm} & 66.27 $\pm$ 6.03 \\ [1ex]
\hline
\end{tabular}
\caption{Comparison of overall accuracy between Random Forest, morphoGNN, and SVarM in road network dataset.}
\label{tab:comparison_accuracy_roads}
\end{table}

For comparison, we also evaluated morphoGNN~\cite{zhu2023data}, a graph neural network model incorporating triplet loss and cross-entropy loss, trained for 50 epochs across 10 independent runs on the same dataset. It achieved overall accuracy of 60.00 $\pm$ 4.46. Additionally, we evaluated SVarM~\cite{hartman2025svarm}, a recently proposed approach leveraging the representation of shapes as varifolds. SVarM achieves an overall accuracy of 66.27 $\pm$ 6.03 on the same dataset. Those comparative results are summarized in Table~\ref{tab:comparison_accuracy_roads}.

\subsection{Neuron trace morphology in mouse brains}
\subsubsection{Data description}
As a second data example, we selected a subset of 3D neuron morphological traces that are part of the NeuroMorpho.org repository (version 8.6.83, released on June 27, 2025), a centrally curated inventory of digitally reconstructed neurons from various species, brain regions, and cell types \cite{Ascoli2007}. For meaningful comparative analysis, data were filtered to include only mouse neurons in five selected brain regions: Basal Ganglia (74 samples), Cerebellum (96 samples), Hippocampus(84 samples), Main Olfactory Bulb(106 samples) and Retina(83 samples).

\begin{figure}[htbp]
  \centering
  \includegraphics[width=0.9\textwidth]{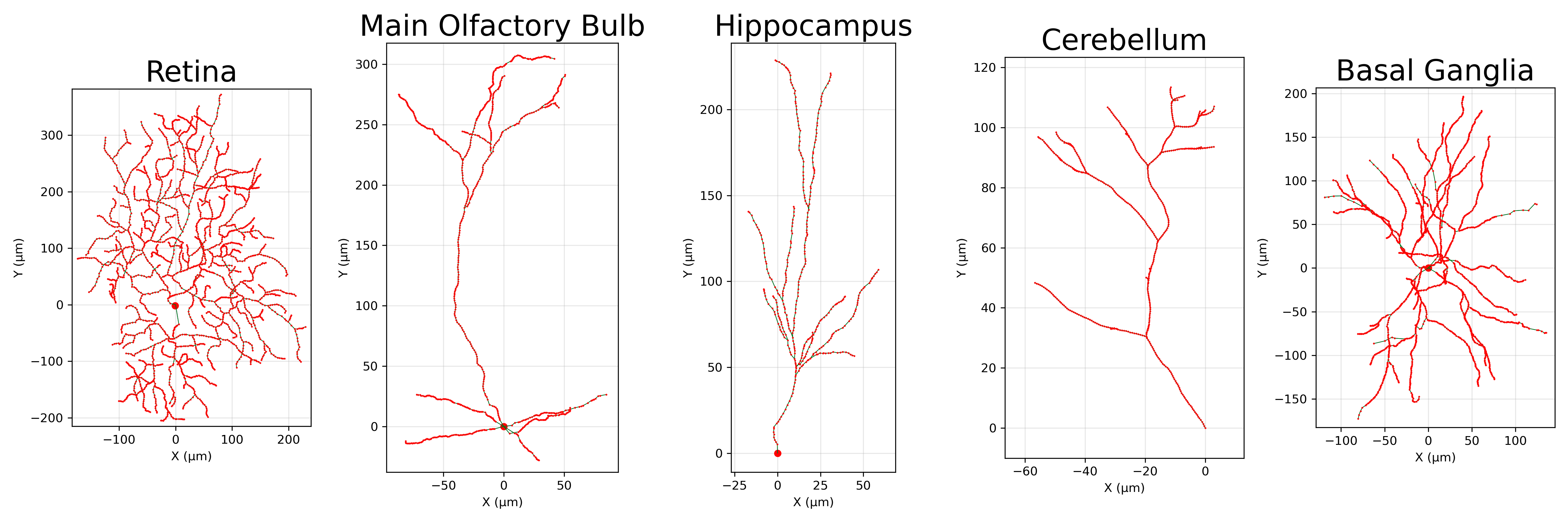} 
  \caption{Visualization of neuron structures from five brain regions in mouse: Retina, Main Olfactory Bulb, Hippocampus, Cerebellum, and Basal Ganglia. The 3D shape graphs are here shown in 2D XY projection to highlight regional morphological differences.}
  \label{fig:neuron_regions}
\end{figure}

Reconstructions are contributed by researchers worldwide, derived from experimental tracings of neuronal morphologies using techniques such as light microscopy and digital reconstruction software. The regions of the brain specifically selected emphasize structural heterogeneity across neural circuits involved in motor control (Basal Ganglia), coordination (Cerebellum), memory (Hippocampus), olfaction (Main Olfactory Bulb), and vision (Retina). To standardize and clean the data, files were loaded using library NetworkX. Region-specific adjustments were applied: in Retina, axons (type=2) were removed, retaining only soma (type=1) and dendrites (types=3,4); other regions did not require removal as they lack axons by default, with Cerebellum additionally lacking soma. Figure \ref{fig:neuron_regions} displays examples of the resulting processed shape graphs for each neuron group.  

\subsubsection{Clustering results}
Similarly to the above dataset, we again compute the set of morphological features that we standardize and evaluate pairwise similarities between neurons as measured by the Euclidean distance in this normalized feature space. The resulting distance matrix is embedded into two dimensions using t-SNE, providing a low-dimensional visualization of the whole dataset. Agglomerative clustering (with 5 clusters) is once again applied to the embedded data points with the cluster quality being evaluated based on the Adjusted Rand Index (ARI) between the unsupervised cluster labels and the known brain-region labels.

\begin{figure}[ht]
    \centering
    \includegraphics[width=1.1\linewidth]{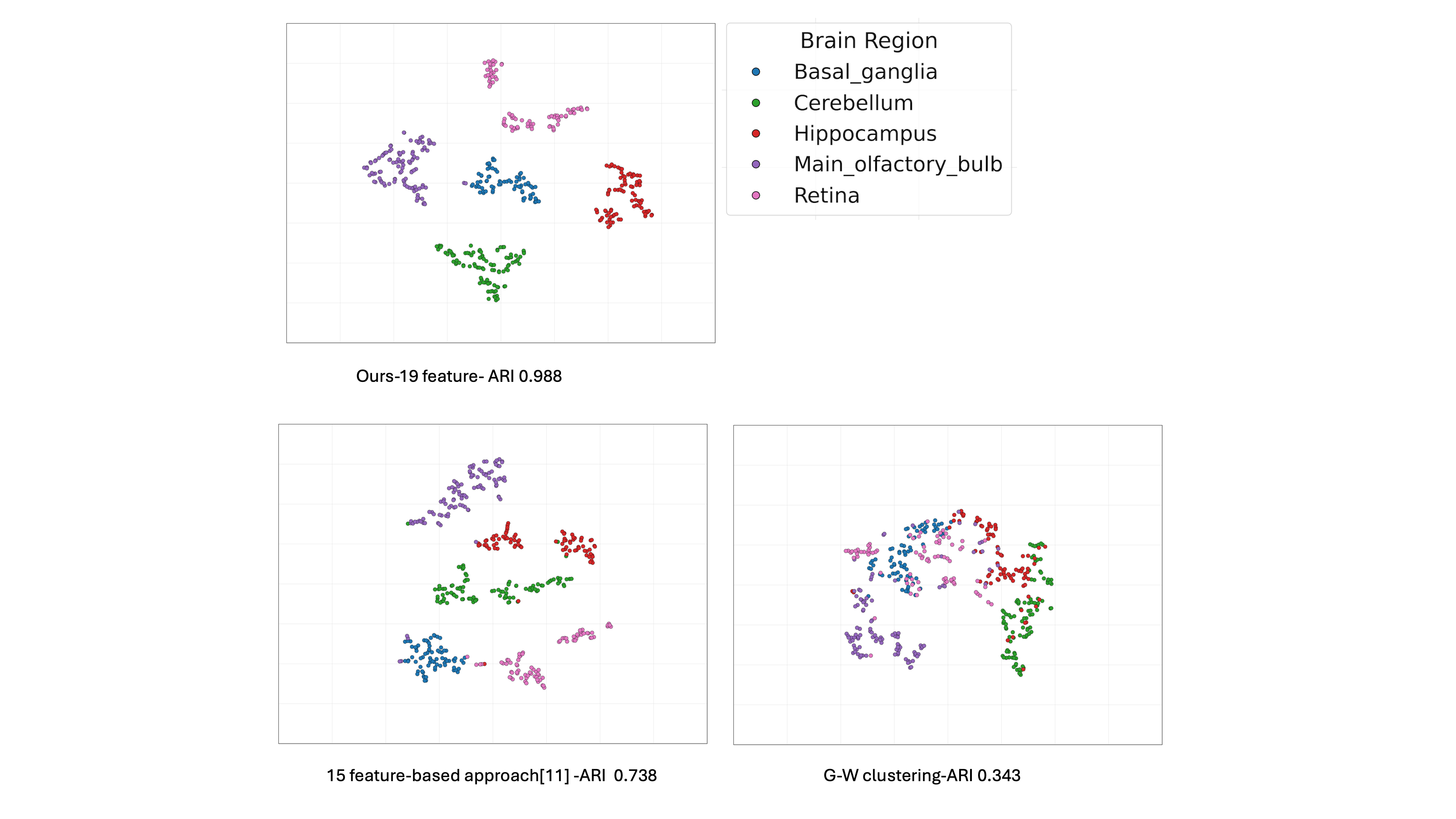}
    \caption{Comparison of clustering for neuronal morphologies obtained from different methods, visualized via t-SNE embedding. Adjusted Rand Index (ARI) between the predicted clusters and ground-truth brain regions are specified below each plot. Our proposed 19-feature representation achieves the highest ARI and yields the clearest visual separation of neurons by region.}
    \label{fig:tsne_neuron_clustering}
\end{figure}

The resulting embedding and ARI is shown on Figure \ref{fig:tsne_neuron_clustering}, which we further compare to the same pipeline applied to two alternative distance representations. First, the same Gromov--Wasserstein distance (G-W) approach as in Section \ref{ssec.cluster_neuron} computed directly from the neuronal tree graphs. Second, we consider the 15-dimensional morphometric feature representation proposed by \cite{Bijari2021} for neuronal shape analysis. We observe that both feature models achieve good and coherent separation between the different brain regions with respective ARI of 0.988 for our feature model and 0.738 for that of \cite{Bijari2021}, against a lower ARI of 0.343 and less clear cluster separation for the G-W approach. Note that we did not include, for this dataset, the group MD matrix described in Section \ref{ssec.group_comp}, as it does not add much more information than the clustering results presented here.

\subsubsection{Classification results}
To further confirm the previous findings, we also evaluate the feature-based classification approach of Section \ref{ssec.classif} on this dataset. We again employ a multiclass Random Forest model within a One-vs-Rest framework trained on the computed features, allocating 70\% of the dataset to training and 30\% to testing and execute 10 random divisions to reduce bias in the reported results. As shown in Table~\ref{tab:neuron_results}, the model is able to very successfully anticipate neurons' brain regions, attaining an overall accuracy of 99.70 $\pm$ 0.37. The low standard deviation also confirms the approach's reliability across different sample choices. Accuracies per region are also reported, indicating consistent performance with slightly greater variability in the retina category. When it comes to the specific distinguishing capabilities of features, we find that the leading features for this dataset, according to the MDI criterion, are circuity, number of edges, average path length, orientation order and  mean betweenness centrality. This is consistent with qualitative visual differences between classes that can be seen in the examples of Figure \ref{fig:neuron_regions}.
\begin{table}[ht]
\begin{minipage}{.45\linewidth}
\centering
\begin{tabular}{l r}
\hline\hline
Brain Region & Accuracy \\ [0.5ex]
\hline
BG & 99.55 $\pm$ 1.36 \\
CB & 100.00 $\pm$ 0.00 \\
HP & 100.00 $\pm$ 0.00 \\
MOB & 100.00 $\pm$ 0.00 \\
RT & 98.80 $\pm$ 1.83 \\
\hline
Overall accuracy & 99.70 $\pm$ 0.37 \\ [1ex]
\hline
\end{tabular}
\end{minipage}
\hfill
\begin{minipage}{.55\linewidth}
\centering
\begin{tabular}{l r}
\hline\hline
Model & Overall Accuracy \\ [0.5ex]
\hline
Random Forest(Ours) & 99.70 $\pm$ 0.37 \\
morphoGNN~\cite{zhu2023data} & 93.26 $\pm$ 2.33 \\
SVarM~\cite{hartman2025svarm} & 97.45 $\pm$ 2.28\\ [1ex]
\hline
\end{tabular}
\end{minipage}
\caption{(Left) Mean accuracy and standard deviation for Random Forest classification of neuron locations in brain regions. (Right) Comparison of overall accuracy between Random Forest, morphoGNN, and SVarM on the neuron dataset.}
\label{tab:neuron_results}
\end{table}

For comparison, we also compare those results to the same two methods as in Section \ref{ssec.classif_roads}, namely morphoGNN~\cite{zhu2023data} and SVarM~\cite{hartman2025svarm}. We see that all three methods achieve high classification scores on this particular dataset, as Table~\ref{tab:neuron_results} (Right) shows, although our proposed feature-based approach still performs slightly better than the other two.   


\subsection{Astrocyte morphology}
\label{ssec.results.astrocyte}

\subsubsection{Dataset description and data preparation}
\begin{figure}[htbp]
  \centering
  \includegraphics[width=1\textwidth]{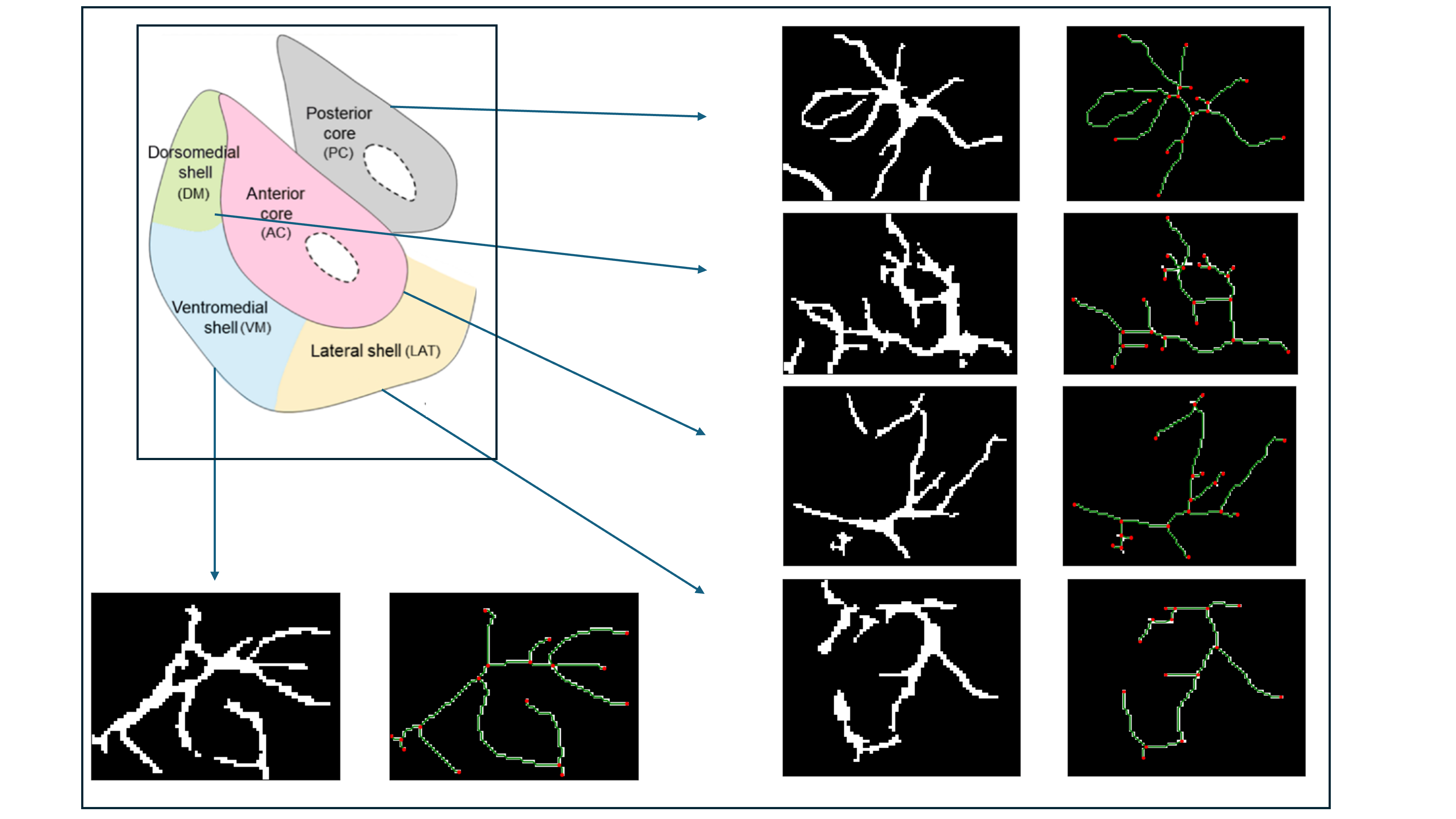} 
  \caption{Representative segmentated astrocytes and the corresponding skeletons computed from images acquired in different subregions of the nucleus accumbens (anatomical map in the top left).}
  \label{fig:skeleton_graph}
\end{figure}

Astrocytes are a  subtype of glial cells found in the brain and spinal cord that play a critical role in the regulation of the central nervous system. One of their remarkable properties is the ability to change shape and size in response to external stimuli or injury (e.g., spinal cord injury); therefore, there is a significant interest in quantifying their shape properties~\cite{Marini2025}. The imaging data considered in this section was originally introduced by one of the authors in \cite{huang2022automated};  we refer to the original article for a detailed description of cellular preparation and image acquisition. The dataset we consider comprises fluorescent GFAP-stained  images of astrocytes sampled from five distinct subregions of the nucleus accumbens in adult rats (cf.~Fig.~\ref{fig:skeleton_graph}): anterior core (AC), posterior core (PC), dorsomedial shell (DM), ventromedial shell (VM), and lateral shell (LAT).  For this study, we considered a total of 2696 single-cell images (taken from the control group of the original dataset) divided into the following number of astrocyte images per subregion: 326 (AC), 1032 (DM), 322 (LAT), 735 (PC), 281 (VM).

Following segmentation of the astrocyte images, we first applied  morphological closing using a disk-shaped structuring element to smooth the images, fill small background holes, and enhance object continuity~\cite{Fisher1997}. Next, we applied  a standard skeletonization procedure consisting of the iterative application of the morphological operators of erosion (controlled shrinking), opening (to preserve connectivity) and thinning (removing boundary pixels). Finally, we discarded all skeleton fragments comprising fewer than 6 pixels to remove noise and irrelevant structures. We subsequently converted the skeletons into graphs, where nodes denote junctions or endpoints, and edges represent the linear paths along the skeleton. Small disconnected graph components were assessed relative to the largest component: if their size ratio fell below 5\%, they were removed. For the remaining small components, we implemented a distance-based reconnection procedure based on the minimum distance between points in the largest and smaller components. If the distance was $\leq$ 6 pixels, a connecting line was drawn between the closest points, merging the smaller component into the largest structure. Components above this distance threshold were discarded. Fig.~\ref{fig:skeleton_graph} shows the segmentation and corresponding skeletons of representative astrocytes extracted from different anatomical regions.

\subsubsection{Astrocyte group comparison}

\begin{figure}[ht]
    \centering
    \includegraphics[width=0.8\linewidth]{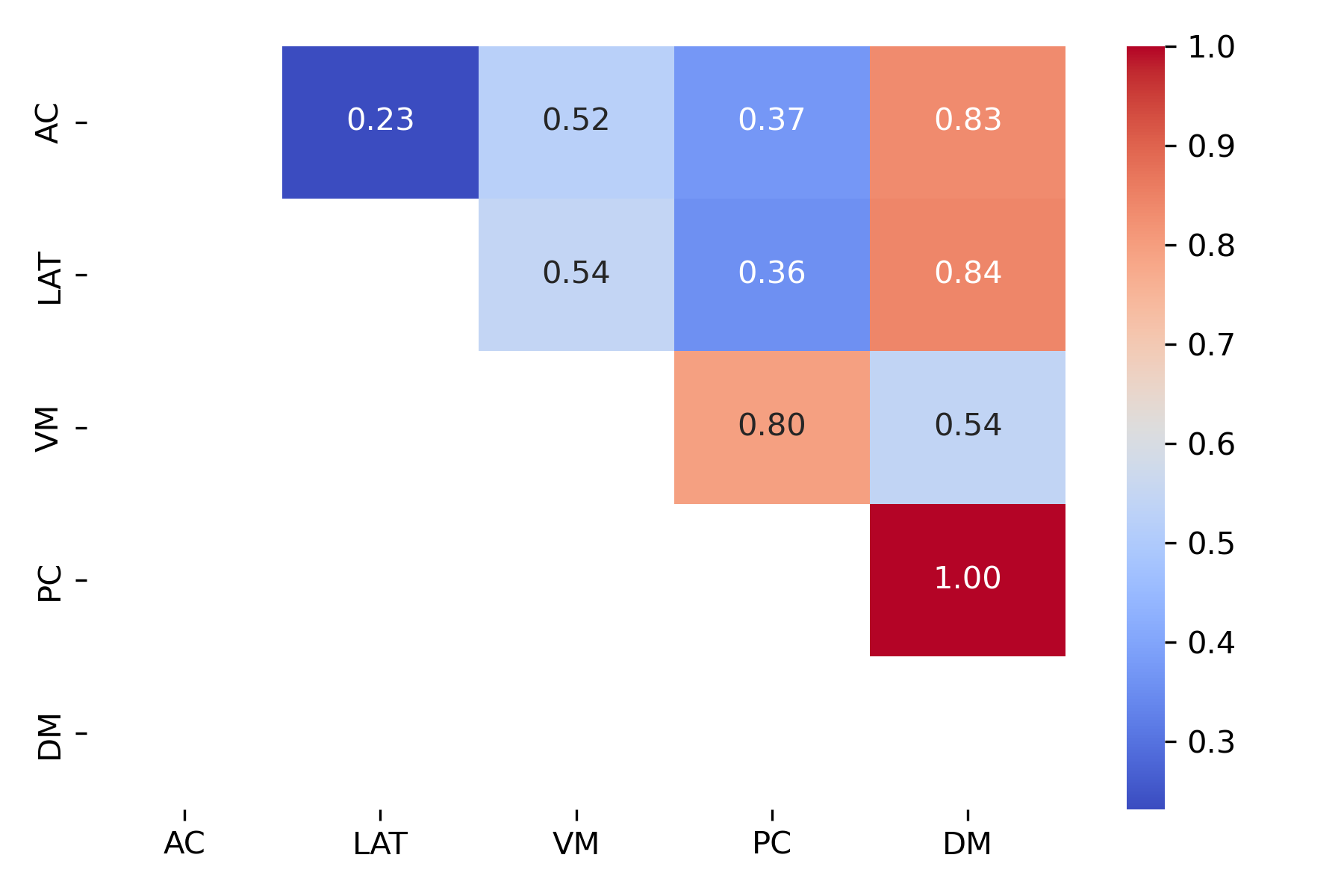}
    \caption{Normalized pairwise morphological distances between the five astrocyte subpopulations associated with the AC, LAT, VM, PC and DM brain subregions.}
    \label{fig:Morphological distance}
\end{figure}

To quantify inter-class structural heterogeneity of astrocyte populations within the NAc, we computed the MD metric derived in eq.~\eqref{eq.aggr_morph_dist} which integrates the 1-Wasserstein distances of the 19 individual morphological features. Figure~\ref{fig:Morphological distance} shows the matrix of the pairwise MD values computed between the populations of the 5 NAc subregions, normalized relative to the maximal observed value. The matrix shows that astrocyte feature distributions in AC are closest to those in LAT, followed by PC, consistent with moderate feature-level distances observed across topological and geometric characteristics. The greatest separation appears between PC and DM, confirming DM's role as the most morphologically distinct subregion. Overall, DM exhibits the most distinctive astrocyte morphological profile compared to other NAc subregions, while AC and LAT demonstrate the highest degree of morphological similarity, reflecting their consistent proximity across individual feature comparisons.

Our result aligns closely with the analysis in~\cite{Marini2025}, which employed the same MD metric but utilized a distinct set of 15 purely geometric features (fractal dimension is the only feature shared in both methods). Both studies identified AC and LAT as having the highest similarity, and both found PC and DM to be the most dissimilar, with DM emerging as the most morphologically distinct region. There are some minor differences in the relative distances between the other 
regions.

While our conclusions align with previous findings, our approach offers complementary insights into the structural landscape of NAc subregions. By incorporating topological features, we reveal a type of structural variation that was not previously observed. Specifically, DM exhibits the most pronounced divergence across topological metrics that include betweenness centrality, assortativity, spectral entropy, and algebraic connectivity. Furthermore, network connectivity measures - most notably graph diameter -  uncover moderate to strong distributional differences, with DM and VM standing out as the most distinct regions.

\subsubsection{Astrocyte Classification results}

We extended the evaluation of the feature-based classification framework to the astrocyte dataset to assess its ability to localize cells within the five NAc subregions (AC, PC, DM, VM, LAT) detailed in Section~\ref{ssec.results.astrocyte}. Adopting the training protocol established in the previous section—a multiclass Random Forest with a One-vs-Rest strategy, utilizing a 70\%/30\% data split averaged over 10 random seeds—we obtained the results summarized in Table~\ref{tab:astrocyte_results}.

The model successfully predicted astrocyte location with an overall accuracy of $76.75 \pm 0.84$. While this performance confirms the discriminative utility of the topological features, the lower accuracy compared to the neuron dataset suggests a higher degree of structural homogeneity or overlap among astrocyte populations across subregions. As detailed in Table~\ref{tab:astrocyte_results} (Left), the classifier proved most robust in the AC and DM regions. Conversely, the LAT and PC subregions exhibited slightly higher error rates, implying greater morphological diversity or class ambiguity in these areas.

To identify the geometric properties driving our classification, we ranked features using the Mean Decrease in Impurity (MDI) criterion. This analysis identified average branch length, algebraic connectivity, circuity, mean betweenness centrality, spectral entropy, and branch count as the primary predictors. These findings corroborate the work in~~\cite{Marini2025}, who similarly underscored the importance of geometric descriptors like sphericity and fractal dimension for astrocyte spatial prediction. Crucially, our approach offers a complementary perspective by revealing the previously unobserved significance of topological features.    

Finally, we benchmarked the proposed approach against morphoGNN~\cite{zhu2023data} and SVarM~\cite{hartman2025svarm}, described in earlier sections. Unlike the neuron classification task where all models performed comparably, the astrocyte dataset revealed a distinct performance gap (Table~\ref{tab:astrocyte_results}, Right). The morphoGNN model achieved an accuracy of $53.21 \pm 1.38$, while SVarM reached $62.68 \pm 1.32$. The more substantial performance degradation observed in these baseline methods likely stems from a lack of rigid motion invariance which is particularly impactful for the astrocyte dataset, where the absence of a natural structural alignment makes robustness to rotation and translation essential.. 

\begin{table}[ht]
\begin{minipage}{.45\linewidth}
\centering
\begin{tabular}{l r}
\hline\hline
NAc Location & Accuracy \\ [0.5ex]
\hline
AC & 79.74 $\pm$ 1.64 \\
DM & 78.70 $\pm$ 1.84 \\
LAT & 74.32 $\pm$ 1.79 \\
VM & 76.96 $\pm$ 1.94 \\
PC & 74.02 $\pm$ 2.90 \\
\hline
Overall & 76.75 $\pm$ 0.84 \\ [1ex]
\hline
\end{tabular}
\end{minipage}
\hfill
\begin{minipage}{.55\linewidth}
\centering
\begin{tabular}{l r}
\hline\hline
Model & Overall Accuracy \\ [0.5ex]
\hline
Random Forest (Ours) & 76.75 $\pm$ 0.84 \\
morphoGNN~\cite{zhu2023data} & 53.21 $\pm$ 1.38 \\
SVarM~\cite{hartman2025svarm} & 62.68 $\pm$ 1.32\\ [1ex]
\hline
\end{tabular}
\end{minipage}
\caption{(Left) Mean accuracy and standard deviation for Random Forest classification of astrocyte Locations. (Right) Comparison of overall accuracy between Random Forest, morphoGNN, and SVarM on the astrocyte dataset.}
\label{tab:astrocyte_results}
\end{table}

\section{Conclusion}\label{sec12}
In this paper, we proposed a framework for the statistical analysis of shape graph datasets based on a set of explicitly defined, invariant morphological and topological features. To demonstrate its versatility, we performed a comprehensive evaluation and benchmarking of this approach on datasets originating from three distinct application areas. Our results show that this feature-based approach, when coupled with standard machine learning techniques, remains competitive with state-of-the-art models like graph neural network models, at least within the data-constrained regimes typical of these domains considered here. Our limited set of handcrafted features offers superior interpretability compared  to the more abstract latent/feature representation typically generated by deep learning architectures.

We are aware of the potential limitations of our proposed model, particularly in scenarios involving significantly larger datasets or subtle morphological differences across data that may exceed the capacity of our current feature space. However, a key advantage of our approach is the relative ease with which our initial feature set can be augmented or integrated with existing models to achieve a more granular representation. One promising future direction is to interface our framework with the types of feature representations that appear in the field of topological data analysis, specifically by incorporating persistence homology signatures. Ultimately, we emphasize that the integration of geometric invariance remains a cornerstone of our work, as our experimental results demonstrate its necessity for developing generalizable models in downstream analysis.


\backmatter


\section*{Acknowledgements}
N. Charon and M. Hossen were partially funded by NSF grants DMS-2438562 and DMS-2526631; D. Labate acknowledges the partial support of the Simons Foundation grant MPS-TSM-00002738. The authors thank Mao Nishino and Michela Marini for their help with some of the datasets used in our simulations. 


\begin{appendices}
\section{Directional standard deviation in 3D}\label{secA1}
For the sake of completeness in the presentation, we here provide the details related to the computation of the mean and standard deviation of an unoriented directional histogram in the case $d=3$, since Section \ref{ssec:direc_features} only focused on the 2D case. 

To that end, we will first introduce a metric on $\mathbb{PR}^2$. Although several possible constructions exist, here we select for simplicity a metric that will lead to closed form expressions for the mean and standard deviation, similar to the 2D setting. The idea is to rely on the embedding of $\mathbb{PR}^2$ into the space of $3 \times 3$ real matrices, which associate to any $u \in \mathbb{PR}^2$ the projection matrix $\Pi_u = u u^T$. We note that $\Pi_u$ is indeed independent of the choice of unit vector $\pm u$ and thus $u \mapsto \Pi_u$ is a well-defined and injective map from $\mathbb{PR}^2$ to $\mathbb{R}^{3\times 3}$. We can then simply consider the metric induced by the (rescaled) Frobenius norm on $\mathbb{R}^{3\times 3}$ and define $d(u,v)^2 = \frac{1}{2}\|\Pi_u - \Pi_v\|_F^2$ for any $u,v \in \mathbb{PR}^2$. A simple computation leads to the following alternative expression:
\begin{equation}
\label{eq:dist_proj_space}
    d(u,v)^2 = 1 - (u^Tv)^2 = \sin^2 \theta_{u,v}
\end{equation}
where $\theta_{u,v}$ is the angle between the two unit vectors $u$ and $v$. This distance is again independent of the choice of orientation vector $\pm u$ and $\pm v$ for the two directions. 

Now, we can define the mean direction of the distribution $\mu_{\mathcal{G}}$ as the minimizer:
\begin{equation*}
    \bar{u} = \text{argmin}_{u \in \mathbb{PR}^2} \, \sum_{e,k} s_k^e \, d(u,u_k^e)^2 
\end{equation*}
which is consistent with the construction of weighted averages in Riemannian manifolds. Moreover, one can take the square root of the minimal value above as a measure of standard deviation for $\mu_{\mathcal{G}}$. With the choice of distance in \eqref{eq:dist_proj_space}, this problem can be reframed as the following constrained maximization:
\begin{equation*}
    \text{maximize} \ \ \sum_{e,k} s_k^e \, (u^T u_k^e)^2 = u^T M u \ \ \text{subj. to } \|u\|=1
\end{equation*}
where $M = \sum_{e,k} s_k^e \, u_k^e (u_k^e)^T$ is a symmetric $3 \times 3$ positive semidefinite matrix. The solution is simply given by a unit norm eigenvector associated to the largest eigenvalue of $M$ (that we write $v_{\text{max}}$ and $\lambda_{\text{max}}(M)$, respectively). Thus, we arrive at the following expressions of the average direction and standard deviation of the shape graph's directional distribution:
\begin{equation}
    \label{eq:mean_std_3D}
    \bar{u} = \pm v_{\text{max}}, \ \ \sigma = \sqrt{\text{Tr}(M) - \lambda_{\text{max}}(M)}. 
\end{equation}
where, in the second equality, we used the fact that $\sum_{e,k} s_k^e = \text{Tr}(M)$ as each vector $u_k^e$ is unit norm. We note that the mean direction $\bar{u}$ is uniquely defined in $\mathbb{PR}^2$ as long as the largest eigenvalue of $M$ has multiplicity $1$, which is the case in generic situations. 


\section{Spearman correlation matrices for computed features}\label{secA2}
\begin{figure}[ht]
    \centering
    \includegraphics[width=1.0\linewidth]{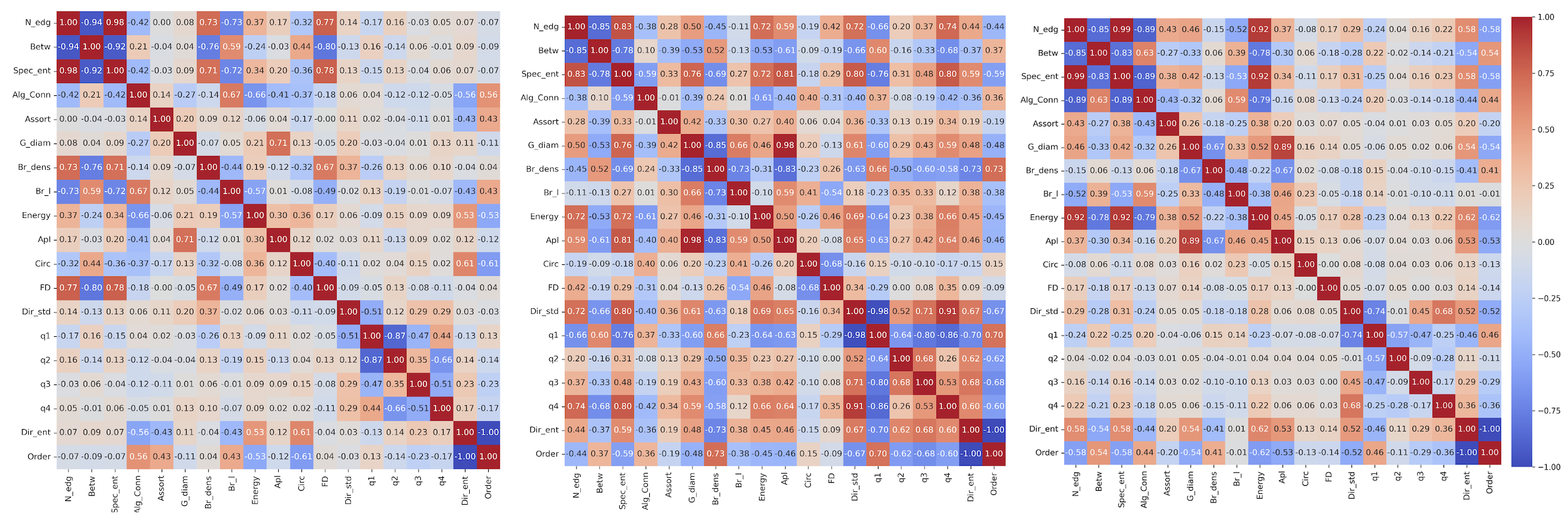}
    \caption{Spearman correlation matrices for the 19 extracted features on the Road network, Neuron traces and Astrocyte datasets.}
    \label{fig:Spearman}
\end{figure}

Figure \ref{fig:Spearman} displays the empirical Spearman correlation matrices $R$ used to compute the aggregated group MD in \eqref{eq.aggr_morph_dist}, for each of the three dataset considered in this work. 

\end{appendices}

Aside from the computation of \eqref{eq.aggr_morph_dist}, these correlation matrices can be also used to reveal interesting relationship patterns or as a measure of redundancy between the different features. For example, as a sanity check, we observe a perfect anticorrelation, across all three datasets, between the directional entropy and orientation-order feature, which is expected as the latter is an explicit inverse function of the former. In the third dataset (astrocyte), we notice a much stronger correlation of $0.92$ between the number of edges and the bending energy as well as between the spectral entropy and bending energy; this is consistent with the observed properties of this dataset, in which shape graphs with larger number of edges tend to have more curved branches and complex connectivity structure.

\newpage
\bibliography{main_draft} 

\end{document}